\documentclass[10pt,twocolumn]{article} 
\usepackage{simpleConference}
\usepackage{times}
\usepackage{graphicx}
\usepackage{amssymb}
\usepackage{floatrow}
\usepackage{multirow}

\usepackage{times}
\usepackage{epsfig}
\usepackage{graphicx}
\usepackage{amsmath}
\usepackage{amssymb}
\usepackage{comment}
\usepackage{color}

\usepackage{placeins}
\usepackage{subcaption}
\usepackage{ulem}

\usepackage{url,hyperref}
\usepackage{comment}
\begin{document}

\title{All4One: Symbiotic Neighbour Contrastive Learning via Self-Attention and Redundancy Reduction}

\author{
Imanol G. Estepa, 
Ignacio Sarasúa, 
Bhalaji Nagarajan, and 
Petia Radeva\\
}
\maketitle
\thispagestyle{empty}

\begin{abstract}
Nearest neighbour based methods have proved to be one of the most successful self-supervised learning (SSL) approaches due to their high generalization capabilities. However, their computational efficiency decreases when more than one neighbour is used.
In this paper, we propose a novel contrastive SSL approach, which we call All4One, that reduces the distance between neighbour representations using "centroids" created through a self-attention mechanism. We use a Centroid Contrasting objective along with single Neighbour Contrasting and Feature Contrasting objectives. Centroids help in learning contextual information from multiple neighbours whereas the neighbour contrast enables learning representations directly from the neighbours and the feature contrast allows learning representations unique to the features.
This combination enables All4One to outperform popular instance discrimination approaches by more than 1\% on linear classification evaluation for popular benchmark datasets and obtains state-of-the-art (SoTA) results. 
Finally, we show that All4One is robust towards embedding dimensionalities and augmentations, surpassing NNCLR and Barlow Twins by more than 5\% on low dimensionality and weak augmentation settings.
The source code would be made available soon.

\end{abstract}

\section{Introduction}

\begin{figure}[h!]
\centering
\includegraphics[width=1.0\linewidth]{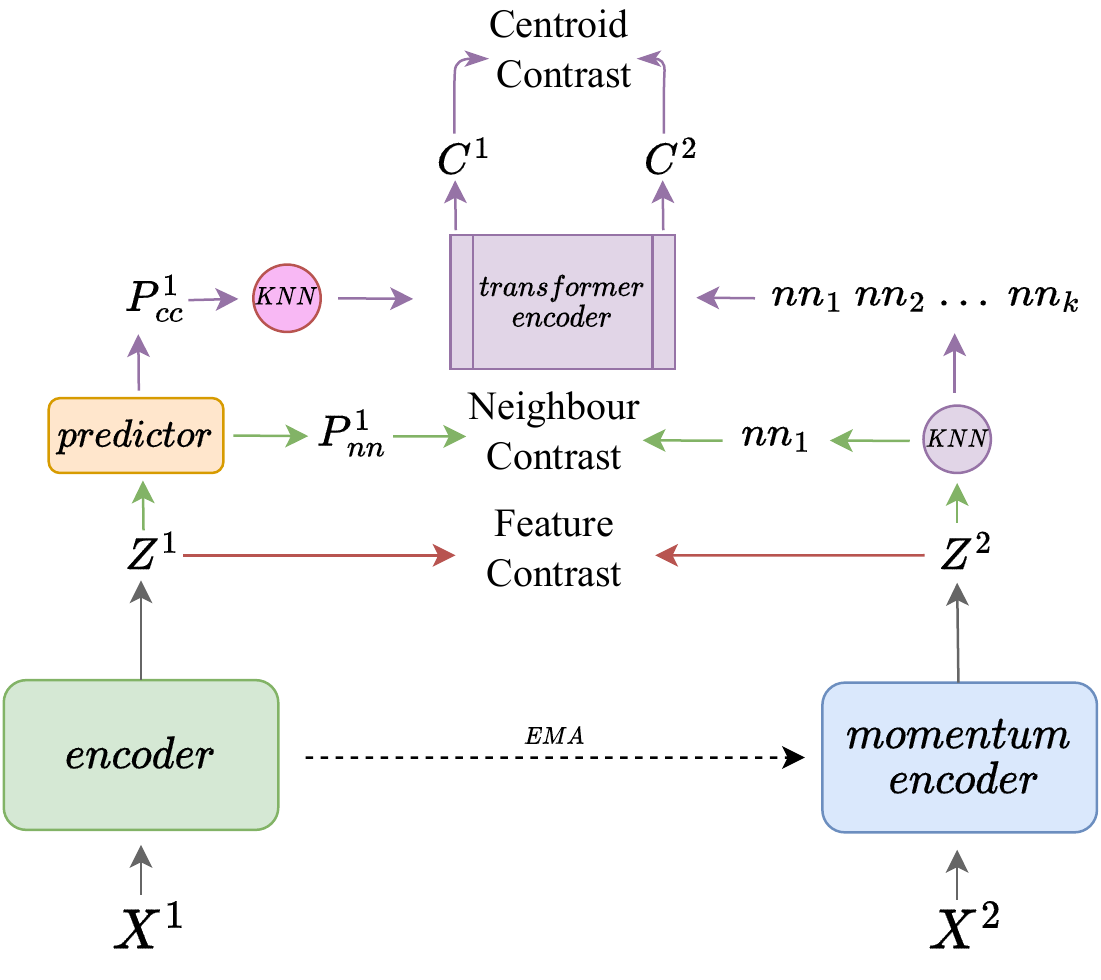}
\caption{\textbf{Simplified architecture of All4One.} All4One uses three different objective functions that contrast different representations:
Centroid objective contrasts the contextual information extracted from multiple neighbours while the Neighbour objective assures diversity \cite{dwibedi_little_2021}. Additionally, the Feature contrast objective
measures the correlation of the generated features and increases their independence.
}
\label{fig:arch}
\end{figure}

Deep learning (DL) models strongly depend on the availability of large and high-quality training datasets whose construction is very expensive \cite{liu_self-supervised_2023}. 
Self-supervised learning (SSL) claims to allow the training of DL models without the need for large annotated data, which serves as a milestone in speeding up the DL progression \cite{liu_self-supervised_2023}.
The most popular SSL approaches rely on instance discrimination learning, a strategy that trains the model to be invariant to the distortions applied to a single image defined as positive samples \cite{chen_simple_2020,he_momentum_2020,tian_contrastive_2020}. 
As all the views belong to the same image, consequently, they belong to the same semantic class. Bringing them together in the same feature space encourages the model to create similar representations for similar images. 
The encouraging results of initial works such as SimCLR \cite{chen_simple_2020} and BYOL \cite{grill_bootstrap_2020} boosted multiple improvements that address common problems of instance discrimination such as lack of diversity between samples and model collapse.

Neighbour contrastive learning hinges on the fact that data augmentations do not provide enough diversity in selecting the positive samples, as all of them are extracted from the same initial image \cite{dwibedi_little_2021}. 
To solve it, Nearest neighbour Contrastive Learning (NNCLR) \cite{dwibedi_little_2021} proposes the use of nearest neighbours (NN) to increase the diversity among the positive samples which in turn boosts the generalization of the model. 
Instead of bringing together two distortions created from the same image, they increase the proximity between a distorted sample and the NN of another distorted sample.
However, relying entirely on the first neighbour holds back the real potential of the approach. 
MSF \cite{koohpayegani_mean_2021} proposes the use of $k$ neighbours to increase the generalization capability of the model. However, MSF suffers from high computation as the objective function needs to be computed for each neighbour ($k$ times). 
Apart from the low diversity of positive samples, instance discrimination approaches suffer from model collapse, a scenario where the model learns a constant trivial solution \cite{grill_bootstrap_2020}. Barlow Twins \cite{zbontar_barlow_2021} proposes a redundancy reduction-based approach that naturally avoids the collapse by measuring the correlation among the features on the generated image representations. 
However, this collapse avoidance suffers from the requirement of projecting embeddings in high dimensions.

In our work, we contrast information from multiple neighbours in a more efficient way 
by avoiding multiple computations of the objective function. This way, we are able to increase the generalization from neighbour contrastive approaches while avoiding their flaws.
For that, we propose the use of a new embedding constructed by a self-attention mechanism, such as a transformer encoder, that combines the extracted neighbour representations in a single representation containing contextual information about all of them. Hence, we are able to contrast all the neighbours' information  on a single objective computation. 
We make use of a Support Set that actively stores the representations computed during the training \cite{dwibedi_little_2021} so that we can extract the required neighbours.
In addition, we integrate our approach with a redundancy reduction approach  \cite{zbontar_barlow_2021}. Making the computed cross-correlation matrix close to the identity reduces the features redundancy of the same image representation while also making them invariant to their distortions. This idea contrasts the representations in a completely different way than the rest of instance discrimination approaches \cite{oord_representation_2019, sohn2016improved,dwibedi_little_2021}.
For this reason, we increase the richness of the representations learnt by the model by combining the 
neighbour contrast approach with the redundancy reduction objective that directly contrasts the features generated by the encoder and aims to increase their independence. 
In addition, the need for high-dimensional embeddings of redundancy reduction feature contrast approaches \cite{zbontar_barlow_2021} is alleviated thanks to our SSL objective combination.

As a summary, in this paper, we introduce a new symbiotic SSL approach, which we call All4One, that leverages the idea of neighbour contrastive learning while combining it with a feature contrast approach (see Figure \ref{fig:arch}). All4One integrates three different objectives that prove to benefit each other and provide better representation
learning. 
Our contributions are as follows: (i) We define a novel objective function, centroid contrast, that is based on a projection of sample neighbours in a new latent space through self-attention mechanisms. (ii) Our proposal, All4One, is based on a combination of centroid contrast, neighbour contrast and feature contrast objectives, going beyond the single neighbour contrast while avoiding multiple computations of the objective function; (iii) We demonstrate how contrasting different representations (neighbours and distorted samples) using InfoNCE \cite{oord_representation_2019} based and feature contrast objectives benefit the overall performance and alleviate individual flaws such as the reliance of high-dimensional embeddings on feature contrast approaches; (iv) We show that All4One, by contrasting the contextual information of the neighbours and multiple representations, outperforms single nearest neighbour SSL on low-augmentation settings and low-data regimes, proving much less reliance on augmentations and increased generalization capability; and (v) We prove that All4One outperforms single neighbour contrastive approaches (among others) by more than 1\% in different public datasets and using different backbones.

\begin{figure}[t]
\begin{center}
\includegraphics[width=1.0\linewidth]{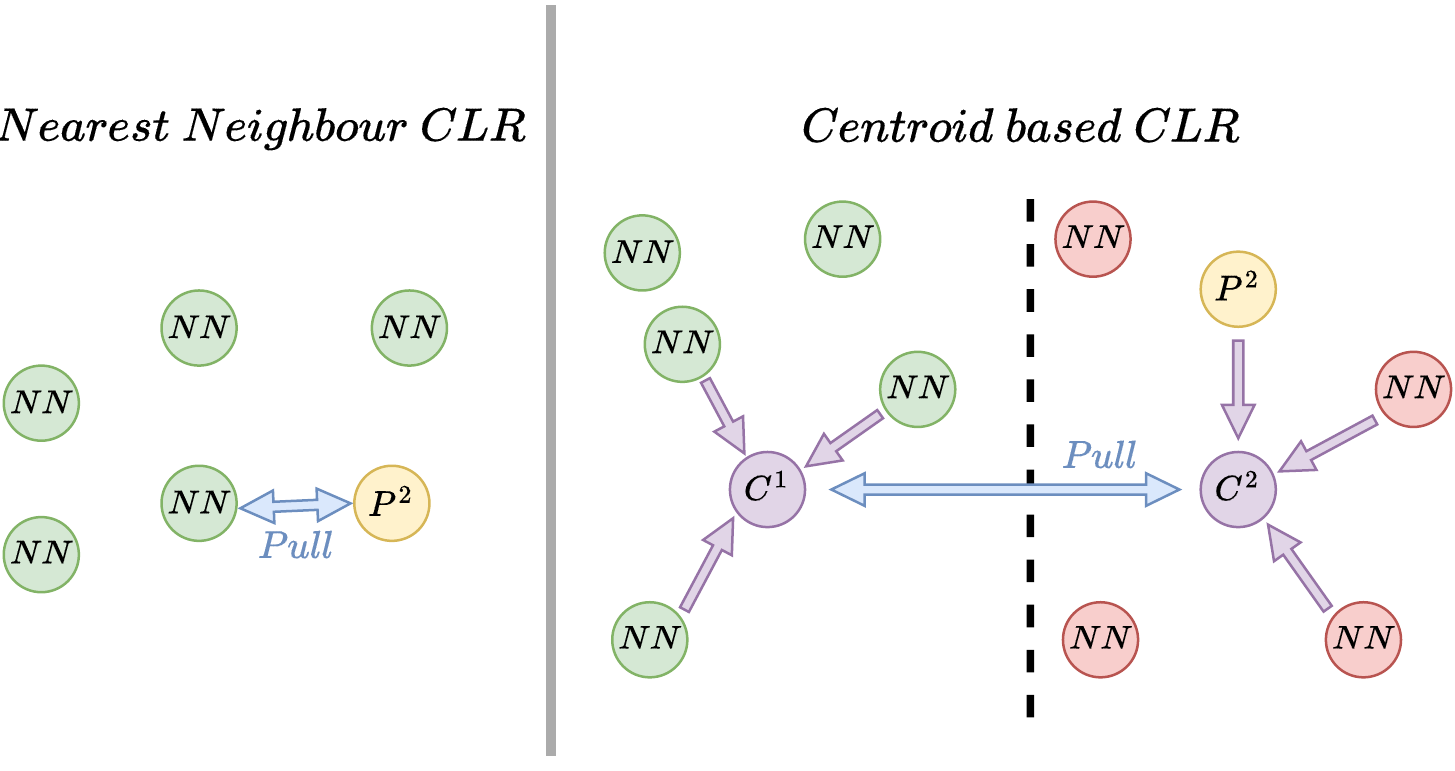}
\end{center}
   \caption{\textbf{Neighbour contrast comparison.} While the common neighbour contrastive approaches only contrast the first neighbour, we create representations that contain contextual information from the \textit{k} NNs and contrast it in a single objective computation.}
\label{fig:short}
\end{figure}

\section{Related Works}
 \textbf{Self-supervised Learning.} 
 In SSL methods, a model is trained to learn intermediate representations of the data in a completely unsupervised way to be transferred later to multiple 
 tasks \cite{liu_self-supervised_2023}. 
 Since the introduction of contrastive loss \cite{chopra_learning_2005}, several works such as SimCLR \cite{chen_simple_2020} and MoCo \cite{he_momentum_2020} proved their usefulness in multiple downstream tasks by proposing variations of InfoNCE \cite{oord_representation_2019}, a contrastive objective function inspired by Noise Contrastive Estimation (NCE) \cite{ma_noise_2018}. In image representation learning, this objective function works with the assumption of positive pairs \((z^a, z^+)\), formed by two representations that share the same semantic class, and negative pairs \((z^a, z^-)\), which are formed by image representations that do not belong to the same semantic class. Thus, the positive pairs are pulled  together in the same feature space while the negative pairs are repelled to avoid  model collapse. Later, BYOL \cite{grill_bootstrap_2020} proved  that it is possible to achieve the same effect without using negative samples by avoiding the collapse with the introduction of architectural changes such as a predictor. Additionally, Barlow Twins \cite{zbontar_barlow_2021} introduced a novel objective function based on the redundancy reduction principle instead of InfoNCE that naturally avoided the collapse.

Overall, discriminative frameworks have been obtaining exceptional results \cite{caron_unsupervised_2020, chen_big_2020, chen_improved_2020, chen_simple_2020} due to the improvements done by introducing new architectures \cite{he_momentum_2020, caron_unsupervised_2020, chen_exploring_2021}, applying new objective functions individually (e. g. redundancy reduction) \cite{noroozi_boosting_2018, barlow_redundancy_2001, ballus_opt-ssl_2022}, alternative augmentation settings \cite{tomasev2022pushing} or even proposing novel strategies such as NN based approaches \cite{dwibedi_little_2021, koohpayegani_mean_2021}, providing an increased generalization by contrasting distortions with NNs. 

During  training, NN based approaches store the representations in a queue and extract them by applying a $k$-NN algorithm that uses one of the representations in the positive pair as a query. This way, one of the pair representations is swapped by its neighbour for the loss computation. Nevertheless, using a single neighbour per sample holds back the potential of the approach meanwhile using \(k\) neighbours reduces the efficiency of the NN approaches as the objective function needs to be computed multiple times. 

 \textbf{Self-attention.} Since the introduction of the Transformer \cite{vaswani_attention_2017} architecture, self-attention-based models have proved to be one of the most successful approaches in Computer Vision (CV). In fact, multiple works analyse the behaviour of the self-attention mechanism and combine it with other well-known tools such as $k$-NN algorithms \cite{wu_centroid_2021, guo_beyond_2022, kreuzer_rethinking_2021, wang_kvt_2022}.
In SSL, Self-attention has been widely used on generative frameworks \cite{he2022masked, bao2021beit, zhou2021ibot}, where they train the transformer backbone to reconstruct the given masked image. However, these reconstruction objectives used in works such as iBOT \cite{zhou2021ibot}, BEiT \cite{bao2021beit} and MAE \cite{he2022masked} are computationally expensive and rely on vision transformers exclusively. In our research, different to previous MIM works, we still maintain a contrastive objective. 
We take advantage of the capacity of the transformer encoder to mix the neighbour representations into a single one that contains information from the neighbours and contrast it using a variation of the InfoNCE \cite{oord_representation_2019}.

\section{The All4One Symbiosis}

In order to increase the performance and efficiency of previous neighbour contrastive methods, we propose a symbiotic SSL framework that combines three different approaches into one. We show our proposed All4One pipeline in Figure \ref{fig:arch}, where we show the three different objectives: the first objective is a neighbour contrast objective (green path);
the second objective is a centroid contrast objective (purple path), which is carried out by the application of self-attention mechanisms \cite{vaswani_attention_2017} and the final objective is a redundancy reduction based feature contrast objective (red path). 
During training, initial representations are transformed depending on the followed path to adapt them to the objective of the path. 

The pipeline is composed of a pair of encoders or neural networks, \(f\), and a pair of projectors, \(g\). The projector consists of a basic MLP that transforms the output of the encoders \cite{chen_simple_2020}. In this case, we apply a momentum encoder, which is a smoothed version of the online encoder, similar to BYOL \cite{grill_bootstrap_2020}.
The pipeline is iterated by a batch of images, $X$. 
For each image, two batches of augmented/distorted images, \( X^1\) and \(X^2\) are generated using a data augmentation pipeline. 
Then, both distorted images are fed to the encoder and, next, to the projector. This sequence of encoder-projector is defined as momentum or online branch depending on the encoder used. We call the momentum branch the encoder-projector sequence that contains a momentum projector and vice-versa \cite{grill_bootstrap_2020}.
The output of momentum and online branches can be defined as \(Z^1 = g^\xi(f^\xi(X^1))\) and \(Z^2 = g^\theta(f^\theta(X^2))\) respectively, being \(g\) and \(f\) the projectors and encoders of each branch. Next, we brief each of the objectives used in the proposal.

\subsection{Neighbour Contrast}

NNCLR \cite{dwibedi_little_2021} is the most popular neighbour contrastive approach. Instead of contrasting two distortions of the same image, it uses the simple KNN operator to extract from a queue or Support Set \cite{dwibedi_little_2021} the NN of the first distortion and contrast it against the second distortion using a variant of the InfoNCE \cite{oord_representation_2019}.
For each \(i\)-th pair in the batch, the neighbour contrast loss is defined as:

\begin{equation}
L_i^{NNCLR} = -log (\frac{exp(nn^1_i \cdot p^2_i / \tau )}{\sum_{k = 1}^{N}exp(nn^1_i \cdot p^2_k / \tau )})
\end{equation}\label{nnclrloss}
where $nn_i$ is the $i$-th NN of the representation $z$, $p_i$ is the second distortion, $\tau$ refers to the temperature constant and $N$ is the number of samples in the batch. 
This way, it increases the generalization of the model through the use of more diverse samples. Note that, in our case, the second predictor \(q^c_\theta\) is used for the Neighbour Contrast objective. 

\subsection{Centroid Contrast}

According to MSF \cite{koohpayegani_mean_2021}, using a single NN could be holding back the potential of the approach. 
In fact, they showed that contrasting the second distorted image with multiple NN could provide a better SSL framework that also obtains higher accuracy regarding the selection of the neighbours. 
Nevertheless, contrasting multiple neighbours hurts the computational efficiency of the model, as they need to compute the objective function \(k\) times, where \(k\) is the number of extracted neighbours. 
For this reason, the improvement is severely constrained by computational resources.
Following the idea of using multiple neighbours, we introduce an alternative proposal that does not require multiple-loss computations. 
We compile the relevant information from the extracted \(k\) neighbours to create a pair of representations, defined as "centroids" that contain contextual information about all the neighbours and pull them together in the feature space applying a variation of InfoNCE \cite{oord_representation_2019} objective function. 
This way the generalization of the model is improved without contrasting multiple neighbours one by one.

Once \(Z^1\) and \(Z^2\) are computed from the pair of distorted images, we calculate the cosine similarity between each \(Z\) (query representation) and the Support set \(Q\), a queue that stores the computed representations  \cite{dwibedi_little_2021}. 
Next, we extract a sequence with the \(K\) most similar representations (\(nn^1_{i} = KNN(z^1_i, Q)\)) for each representation in both batches. Then, for each \(Z^1\) and \(Z^2\), we obtain their respective batch of sequences of NNs, \(NN^1\) and \(NN^2\).
As we try to avoid contrasting the neighbours one by one, we introduce a new element to the pipeline: a transformer encoder, \(\psi\). 
Given \(NN^1\) and \(NN^2\), we input \(\psi\) with each sequence \(nn^1_i\) to compute the self-attention of the sequences. 

Given a sequence of neighbour representations \(nn^1_i\), we 
obtain a single representation \(c_1\) that contains as much information as possible about the input sequence \(nn^1_i\). 
When computing self-attention \cite{vaswani_attention_2017}, we mix the representations of the input sequence in a weighted manner so that a new enriched vector of representations is returned. 
Each element of this enriched vector contains contextual information about all the neighbours in the sequence.
During training, for each sequence in \(NN^1\), the process is made up of the following steps: (i) for each sequence \(Seq_i\) in \(NN^1\), we add sinusoidal positional encoding \cite{vaswani_attention_2017}; (ii) then, we feed the transformer encoder \(\psi\) with \(Seq_i\); (iii) inside the transformer encoder, self-attention is computed and a new sequence is returned \(Seq^c_i\); (iv) finally, we select the first representation \(Seq^c_{i \, 1}\) in the returned sequence \(Seq^c_i\) as our centroid \(c_i\) as we aim to contrast a single representation that contains context information from the rest of the neighbours. After selecting the first representation on all sequences, we obtain a batch of representations defined as \(C^1\).

On the online branch, a slightly different process is followed. A second MLP similar to the projector, the predictor, is used to change the feature space of \(Z^2\) batch and  obtain \(P^{c2} = q_c(Z^2)\) batch of transformed representations. More concretely, we pass \(Z^2\) through  the centroid predictor \(q_c\). Then, we replace the fifth neighbour in each sequence \( nn^2_{i} \) of \(NN^2\) by \( p^{c2}_i \). Finally, we reorder each sequence so \(p^{c2}_i\) is the first element. We define this process as \(Shift\) operation. This is done to introduce the distorted image in the sequence, thereby impacting the back-propagation. Once modified \(NN^2\) is created, we pass it through the transformer encoder to obtain \(C^2\) by following the previously explained process.
Finally, we contrast \(C^1\) and \(C^2\) using a variation of InfoNCE \cite{oord_representation_2019} loss function aiming to bring the neighbour centroids together (see Figure \ref{fig:short}). 
For each centroid pair, the centroid loss can be defined as:
\begin{equation}
L_i^{centroid} = -log \left( \frac{exp(c^1_i \cdot c^2_i / \tau )}{\sum_{n = 1}^{n}exp(c^1_i \cdot c^2_n / \tau )} \right)
\end{equation}

\subsection{Reducing Redundancy: Feature Contrast}

 The application of the redundancy reduction principle to increase the independence of the features is one of the most successful approaches in the SSL SoTA \cite{zbontar_barlow_2021}. The Barlow Twins' main idea is that, instead of focusing on the images, to directly contrast the features by computing a cross-correlation matrix \(C_{ij}\). Then, it aims to increase the invariance of the features by equating the diagonal elements \(c_{ii}\) of \(C_{ij}\) to one (invariance term) while also decreasing the redundancy between the features by reducing the correlation between different features (\(i^{th}\) and \(j^{th}\) features). Inspired by this approach, we increase the richness of our framework by introducing a feature contrast objective function that measures the correlation of the features and aims to increase their independence. 

To do so, we construct the output of the momentum and online projectors as \((Z)_{ij}\) matrices where each element represents an exact feature \(j\) of a single augmented image representation, \(i\). 
Then, we use \( L_2 \) normalization on both matrices in the batch dimension \cite{zbontar_barlow_2021} and we compute the cosine similarity between the transposed matrix, \(Z^{1T}\) and \(Z^{2}\) to obtain the cross-correlation matrix, \(CC^1\). We compute this term symmetrically. This is done by 
by swapping the branches and computing the similarities. 
Finally, two cross-correlation matrices \(CC^1\) and \(CC^2\) are obtained. Redundancy reduction feature contrast objective \cite{ballus_opt-ssl_2022} is computed as follows:
\begin{equation}
\begin{aligned}
L_{Red.} = \sqrt{\frac{1}{2D}\sum_{i=1}^{D}((1 - cc^1_{ii})^2 + (1 - cc^2_{ii})^2)}\\
+ \lambda \sqrt{\frac{1}{2D(D-1)}\sum_{i=1}^{D}\sum_{j \neq i}^{D} ((cc^1_{ij})^2 + (cc^2_{ij})^2)}
\end{aligned}
\end{equation}

As it can be noted, the first term increases the correlation between the elements that represent the same feature among the distorted images (diagonal elements) while the second term decreases the correlation between elements that represent different features (off-diagonal elements). In our framework, we empirically set  \(\lambda\) to 0.5.

\subsection{Final Objective: The All4One Objective}
Once all objectives are computed, the final loss function is formed 
by summing the previously defined objectives.
The All4One objective is defined as:
\begin{equation} \label{floss}
    L_{All4One} = \sigma L_{NNCLR} + \kappa L_{Centroid} + \eta L_{Red}
\end{equation}
where \(\sigma\), \(\kappa\) and \(\eta\) were determined empirically as 0.5, 0.5 and 5, respectively. By combining different objectives, the All4One objective improves the learning of representations. We show the improvements over other methods below.

\section{Experiments}\label{experiments}
In this section, we first describe the implementation details of All4One and its training.
Then, we evaluate it using the common image classification linear evaluation pipeline
on different datasets: CIFAR-10 
 \cite{krizhevsky2009learning}, CIFAR-100 \cite{krizhevsky2009learning}, ImageNet-1K \cite{krizhevsky_imagenet_2012} and ImageNet-100, a reduced ImageNet of 100 classes. We extensively study the different components of our proposal and discuss the various design decisions in detail. Finally, we also extend our proposal to a transformer-based backbone using ViT \cite{dosovitskiy2020image} and also show the efficacy of our proposal on other downstream tasks.

\subsection{Implementation Details}\label{details}
\textbf{Architecture.} All4One follows a momentum instance discrimination pipeline (Figure \ref{fig:arch}). The momentum branch 
 includes the usual momentum projector, \(g_\xi\) \cite{grill_bootstrap_2020}. The online branch, on the contrary, includes a projector, \(g_\theta\) and double predictor, \(q_{nn}\) and \(q_{c}\). MLP projectors are formed by 3 fully connected layers of size [2048, 2048, 256], while the MLP predictor uses 2 fully connected layers with a dimensionality of [4096, 256]. Similar to NNCLR \cite{dwibedi_little_2021}, all fully-connected layers, except the last ones, are followed by batch-normalization \cite{ioffe_batch_2015}.

MLP components of a SSL method 
filter the features generated by the encoder, keeping those that 
are useful for the downstream task \cite{appalaraju_towards_2020}.  In All4One, there are multiple objectives that are applied in completely different representations, so the predictors involved should filter the features independently. For this reason, we propose the use of two different predictors that work separately for each objective.
Finally, our approach introduces a small Transformer encoder, \(\psi\), that is applied for both branches and a shared \(KNN(\cdot,\cdot)\) operator that extracts image representations from the Support Set, \(Q\), given a query.
The final objective of our proposal follows Eq. (\ref{floss}), which brings together both contrasting approaches. More details can be found in Appendix A.

\textbf{Training.} We train All4One on CIFAR-10, CIFAR-100, ImageNet100 (using a ResNet-18 backbone) and the complete ILSVRC2012 ImageNet (using a ResNet-50 backbone) without any class label or annotation.
During the training, all backbones are initialized with default Solo-learn \cite{costa_solo-learn_2022} initialization. All MLP components use the PyTorch default initialization. The transformer encoder uses three transformer encoder layers with 8 heads each \cite{vaswani_attention_2017}.
Following SimCLR \cite{chen_simple_2020}, \(lr\) is adapted using \(lr * batch size /256\) formula and readapted for each layer using LARS \cite{you_large_2017} (only for CNNs). For ResNet-18 and ResNet-50, we use SGD optimizer and on the ViT-Small backbone, we use AdamW. We find it optimal to use 1.0, 0.00015 and 0.1 as base learning rates for the CNN backbones, ViT and the internal transformer encoder respectively, which are gradually reduced by using a warm-up cosine annealing scheduler. When using AdamW, we 
detach the \(lr\) of the transformer encoder by setting it to a constant of 0.1. 
Finally, the queue or Support set size is set to 98304 for fair comparison to NNCLR \cite{dwibedi_little_2021}. The rest of the hyperparameters of the model are directly extracted from NNCLR \cite{dwibedi_little_2021}.
All the training processes are done on a single NVIDIA RTX 3090 GPU, except for ImageNet experiments where the evaluations are done on 4xNVIDIA V100 GPUs.

\subsection{All4One Evaluation}

\begin{table}[t]
\addtolength{\tabcolsep}{-2pt}
\begin{tabular}{ccccccc}
                 & \multicolumn{2}{c}{CIFAR-10}    & \multicolumn{2}{c}{CIFAR100}    & \multicolumn{2}{c}{ImageNet100} \\ \cline{2-7} 
Method           & Top-1          & Top-5          & Top-1          & Top-5          & Top-1          & Top-5          \\ \hline
BYOL             & 92.58          & 99.79          & 70.46          & 91.96          & 80.16          & 95.02          \\
DC V2   & 88.85          & 99.58          & 63.61          & 88.09          & 75.36          & 93.22          \\
DINO             & 89.52          & 99.71          & 66.76          & 90.34          & 74.84          & 92.92          \\
MoCoV2+         & 92.94          & 99.79          & 69.89          & 91.65          & 78.20          & 95.50          \\
MoCoV3          & 93.10          & 99.80          & 68.83          & 90.57          & 80.36          & 95.18          \\
ReSSL            & 90.63          & 99.62          & 65.92          & 89.73          & 76.92          & 94.20          \\
SimCLR           & 90.74          & 99.75          & 65.78          & 89.04          & 77.64          & 94.06          \\
Simsiam          & 90.51          & 99.72          & 66.04          & 89.62          & 74.54          & 93.16          \\
SwAV             & 89.17          & 99.68          & 64.88          & 88.78          & 74.04          & 92.70          \\
VIbCReg          & 91.18          & 99.74          & 67.37          & 90.07          & 79.86          & 94.98          \\
VICReg           & 92.07          & 99.74          & 68.54          & 90.83          & 79.22          & 95.06          \\
W-MSE            & 88.67          & 99.68          & 61.33          & 87.26          & 67.60          & 90.94          \\ \hline
BT     & 92.10          & 99.73          & 70.90          & 91.91          & 80.38          & 95.28          \\

NNCLR            & 91.88          & 99.78          & 69.62          & 91.52          & 79.80          & 95.28          \\ \hline

All4One & \textbf{93.24} & \textbf{99.88} & \textbf{72.17} & \textbf{93.35} & \textbf{81.93} & \textbf{96.23} \\ \hline
\end{tabular}
\caption{
\textbf{Linear evaluation results on CIFAR-10, CIFAR-100 and ImageNet100.} The results are extracted from Solo-learn  Self-supervised learning library \cite{costa_solo-learn_2022}.}
\label{tablon}
\end{table}
\textbf{CIFAR and ImageNet100 Linear Evaluations.}\footnote{Qualitative analysis of the representations is shown in Appendix B.}
We compare our approach against the current SoTA SSL frameworks. 
We first show the evaluations on the CIFAR datasets and the ImageNet-100 dataset.
ImageNet-100 is a reduced ImageNet version with 100 classes and the images are 224x224 (as compared to CIFAR which has  32x32).
We report Top-1 and Top-5 linear accuracies for all the datasets.
As can be seen from Table \ref{tablon}, our approach clearly outperforms the previous SoTA approaches, including the ones that inspired our own approach.
We gain 1.36\%, 2.55\% and 2.13\% over NNCLR on CIFAR-10, CIFAR-100 and ImageNet-100 respectively, and similarly improve by 1.14\%, 1.27\% and 1.55\% over Barlow Twins. This emphasizes the fact that we are able to outperform both the feature contrast approach and the neighbour contrast approach by a considerable margin by combining them and adding the new contrastive centroid loss.

\textbf{Linear Evaluation on ImageNet}. 
In table \ref{full_img}, we compare NNCLR \cite{dwibedi_little_2021} and All4One on the complete ILSVRC2012 ImageNet \cite{deng_imagenet_2009} dataset for linear evaluation task.  
Considering the computational resources for ImageNet, we perform the linear evaluation only for 100 epochs with an effective batch size of 1024. 
We are able to outperform NNCLR (all the hyperparameter settings are maintained from its original paper) on the larger ImageNet \footnote{The training curve comparison is shown in Appendix C.}. This highlights the improvements in the performance of our approach on larger datasets.

\begin{table*}[t]
    \begin{subtable}[t]{.45\linewidth}
    \centering
    \begin{tabular}{@{}llll@{}}
    \hline
       Method          &      Top-1 & Top-5\\ \hline
    NNCLR       & 65.74        &   86.90      \\
    
    All4One  & \textbf{66.60}   &    \textbf{87.51}          \\ \hline
    \end{tabular}
    \caption{\textbf{Linear evaluation on ILSVRC2012 ImageNet.} 
    For NNCLR, all hyperparameters except for the batch size (we use 1024 for both approaches) are the ones recommended in the original paper \cite{dwibedi_little_2021}. 
    }\label{full_img}
    \end{subtable}
    \hspace{0.5cm}
    \begin{subtable}[t]{.45\linewidth}
    \centering
    \begin{tabular}{@{}llll@{}}
    \hline
       Method          &      Top-1 & Top-5\\ \hline
    NNCLR       & 68.55        &   90.94      \\
    
    All4One  & \textbf{69.7}   &    \textbf{91.65}          \\ \hline
    \end{tabular}
    \caption{
    \textbf{Linear evaluation on CIFAR-100 using ViT-Small backbone.} Same hyperparameter settings are used for both methods.
    }\label{trans}
    \end{subtable}
    \begin{subtable}[]{.45\linewidth}
    \begin{tabular}{lllll}
    \cline{2-5}
                     & \multicolumn{2}{l}{ImageNet100} & \multicolumn{2}{l}{ImageNet} \\ \cline{2-5} 
    Method           & 1\%            & 10\%           & 1\%           & 10\%         \\ \cline{1-5} 
    NNCLR            & 54.14          & 75.49          & 37.51         & 58.74        \\
    All4One (Ours) & \textbf{58.73}          & \textbf{76.95}          & \textbf{38.96}         & \textbf{60.14}        \\ \hline
    \end{tabular}
    \caption{\textbf{Semi-supervised learning results (Top-1 linear accuracy) on ImageNet100 and ImageNet.} 
    }\label{semi}        
    \end{subtable}
    \hspace{0.5cm}
    \begin{subtable} [] {.45\linewidth}
        \begin{tabular}{lllll}
               & Food-101       & Caltech-101    & Dogs           & Pets           \\ \hline
NNCLR          & 69.52          & 90.15          & 67.47          & \textbf{83.34} \\
All4One & \textbf{71.16} & \textbf{91.10} & \textbf{68.07} & 81.57 \\ \hline        
\end{tabular}
\caption{\textbf{Transfer learning evaluation (Top-1 linear accuracy). }\phantom{aaaaa}
} 
\label{transfer}
    \end{subtable}

  \caption{\textbf{All4One linear evaluation experiments.}}\label{tables}
\end{table*}

\textbf{Linear evaluation on Transformer Backbone.} 
We study the All4One behaviour on a different backbone by replacing the ResNet with a Transformer backbone. We use a ViT-Small on CIFAR-100 for this study. In Table \ref{trans}, we see how All4One outperforms NNCLR verifying the independence of our approach on backbones. We have an improvement of 1.15\% compared to the ViT version of NNCLR. 

\textbf{Semi-supervised ImageNet100 and ImageNet Evaluations}. 
We perform semi-supervised evaluation following the experiments in NNCLR \cite{dwibedi_little_2021}. We fine-tune the ImageNet pre-trained model (ResNet-50 on 100 epochs) on 1\% and 10\%  subsets of the datasets. The results are presented in Table \ref{semi}. 
As can be seen, All4One generalizes better than NNCLR, outperforming it for both ImageNet100 and ImageNet on 1\% and 10\% subsets of the data.

\textbf{Transfer Learning.} 
Finally, we evaluate All4One and NNCLR on transfer learning downstream tasks for Food101 \cite{kaur2017combining}, Caltech-101 \cite{xu2016simple}, Dogs \cite{zhao2020universal} and Pets \cite{parkhi2012cats} dataset. For all datasets, we freeze the ImageNet pre-trained model (ResNet-50 on 100 epochs) and train a single linear classifier on top of it for 90 epochs on train splits while performing a sweep over the \(lr\) to obtain the best-performing one. Then, we evaluate the performance on the validation split for Food101 dataset and test split for the rest of the datasets. 
The results are shown in Table \ref{transfer}.
As can be seen, All4One outperforms NNCLR in 3 out of 4 datasets, further validating the increased generalization capabilities of All4One.

\subsection{Ablation Study}
First, we show the importance of each objective defined by our approach.
Then, 
we analyse the dimensionality and augmentation robustness of our model. 
Finally, we present some design choices such as the number of layers used by the transformer encoder and the number of extracted neighbours. 
All the ablations are done following exactly the same settings defined in Section \ref{details} if not stated otherwise.

\textbf{Objective Importance.} 
As explained, our approach introduces redundancy reduction and novel neighbour contrast objectives. For this reason, we find it interesting to analyse, one by one, the importance of each of them. 
In Table \ref{ablation1}, we report the performance of each objective of 
the framework. 
In addition, we also study the importance of EMA (\(v2\) vs \(v3\)). 
We see that the addition of EMA boosts the overall performance of the model.

\begin{table*}[t]
\centering
\begin{tabular}{@{}llllllll@{}}
\hline
Method       & NNCLR obj. & Cen. obj. & Feat. obj. & EMA & Top-1 & k-NN Top-1 & NN Top-1 \\ \hline
All4One\(_{v0}\)      & \checkmark          &                &                       &     & 69.62 & 62.16      & 68.8 (77.8*)         \\
All4One\(_{v1}\) &            & \checkmark              &                       &     & 67.4  & 59.61      & 82.8         \\
All4One\(_{v2}\) & \checkmark          & \checkmark              &                       &     & 71.02 & 63.21      & 85.28        \\
All4One\(_{v3}\) & \checkmark          & \checkmark              &                       & \checkmark   & 71.08 & 63.83      & \textbf{86.16}        \\
All4One\(_{v4}\) &            & \checkmark              & \checkmark                     & \checkmark   & 71.31 & 63.72      & 80.6         \\
All4One\(_{v5}\) & \checkmark          &                & \checkmark                     & \checkmark   & 71.64 & 64.58      & 78.8         \\
All4One\(_{v6}\) & \checkmark          & \checkmark              & \checkmark                     & \checkmark   & \textbf{72.17} & \textbf{64.84}      & \textbf{82.16}        \\ \hline
\end{tabular}
\caption{\textbf{All4One objective function ablation study (using CIFAR-100).} 
All4One\(_{v0}\) is equal to vanilla NNCLR. NN retrieval accuracy marked by * represents the NN retrieval obtained by increasing the queue size from 65503 to 98304  \cite{dwibedi_little_2021}}
\label{ablation1}
\end{table*}

\begin{table*}[t]
    \begin{subtable}[t]{0.45\linewidth}
    \centering
    \begin{tabular}{@{}llll@{}}
    \hline
                       & Top-1 & k-NN Top-1\\ \hline
    Barlow Twins (2048)      & 71.21      & 63.11                 \\
    Barlow Twins (256) & 62.14      & 54.64                 \\
    All4One (256) & \textbf{72.17}      & \textbf{64.84}                 \\ \hline
    \end{tabular}
    \caption{\textbf{Dimensionality analysis using CIFAR-100 dataset.}} 
    \label{BTdim}
    \end{subtable}
   \begin{subtable}[t]{0.45\linewidth}
   \centering
    \begin{tabular}{@{}lll@{}}
    \hline
    Method                & Top-1      & k-NN Top-1   \\ \hline
    Barlow Twins    \cite{zbontar_barlow_2021}      & 39.66         & 30.89          \\
    NNCLR      \cite{dwibedi_little_2021}           & 35.39         & 27.64          \\
    \textbf{All4One (Ours)} & \textbf{44.7} & \textbf{33.99} \\ \hline
    \end{tabular}
    \caption{\textbf{Augmentation analysis using CIFAR-100.}}\label{augm}
   \end{subtable}
    \begin{subtable}[]{.45\linewidth}
    \centering
    \begin{tabular}{@{}lll@{}}
    \hline
     Layer number & Top-1 & k-NN Top-1 \\ \hline
    \phantom{aaaaa}3 & 72.17 & 64.84 \\
    \phantom{aaaaa}6 & 71.86 & 64.50 \\
    \phantom{aaaaa}9 & 71.75 & 64.52 \\ \hline
    \end{tabular}
    \caption{\textbf{Number of transformer layers.}}
    \label{tab:encoder_study}
    \end{subtable}
    \begin{subtable}[]{.45\linewidth}
    \centering
    \begin{tabular}{@{}lll@{}}
    \hline
     Number of NN  & Top-1 & k-NN Top-1 \\ \hline
    \phantom{aaaaa}5 & 72.17 & 64.84 \\
    \phantom{aaaaa}10 & 72.00 & 64.54 \\
    \phantom{aaaaa}15 & 71.92 & 64.63 \\
    \phantom{aaaaa}20 & 71.79 & 64.6 \\ \hline
    \end{tabular}
    \caption{\textbf{Number of NNs extracted.}}
    \label{tab:neighbour_study}
    \end{subtable}
    \vspace{0.5cm}
  \caption{\textbf{All4One ablation experiments.} Evaluated on CIFAR-100 for linear and k-NN classification.}\label{tables}
\vspace*{-5mm}
\end{table*}

As can be seen with the different versions of All4One, all three objectives are important regarding the overall performance. 
Intuitively, both NNCLR \cite{dwibedi_little_2021} and Centroid-based objectives focus on contrasting image representations, so it is possible that, during the training, both objectives may partially overlap. 
This is not the case for the redundancy reduction, as it focuses on the features instead, causing its removal more critical than the others. 
Moreover, we designed the redundancy objective to use the representations \(z^1_i\) and \(z^2_i\) to compute the loss rather than using the neighbours. This fact adds more richness to the final loss function, as a total of three different representations (original neighbour \(nn^1_i\), centroid derived from the neighbours \(c^1_i\) and image representation \(z^1_i\)) are used for the unified objective. 

On the other hand, we check that the Centroid objective is the one that increases the most on the NN retrieval accuracy, reaching 86.16\% when combined with the NNCLR \cite{dwibedi_little_2021} and EMA architecture (experiment \(v3\)). 
As expected, contrasting contextual information from multiple neighbours encourages the model to create better representations easier to distinguish just by using a simple KNN operator.

\textbf{Dimensionality Robustness.} 
Redundancy reduction approaches such as Barlow Twins \cite{zbontar_barlow_2021} highly depend on the dimensionality of the embeddings.
Other approaches such as Opt-SSL \cite{ballus_opt-ssl_2022} required high dimensional embeddings to provide SOTA results. 
Our approach, however, manages to outperform Barlow Twins with much lower dimensional embeddings as it does not only depend on the redundancy reduction. 
Even if the dimensionality of the embedding is low, the symbiosis formed avoids the decrease in performance, as can be seen in Table \ref{BTdim}.
Another factor of consideration to use low dimensions is that \(KNN(\cdot,\cdot)\) suffers from the curse of dimensionality, which decreases its efficiency when high dimensional embeddings are used.

\textbf{Augmentation Dependency.} 
SSL frameworks depend heavily on the augmentations used by the Pretext task generator to create hard positive samples. However, neighbour contrastive approaches \cite{dwibedi_little_2021, koohpayegani_mean_2021} empirically proved that they are naturally more robust to augmentation removals. 
NNCLR \cite{dwibedi_little_2021} proposes the removal of all the augmentations except random crop augmentation to compare NNCLR frameworks robustness with SimCLR \cite{chen_simple_2020} and BYOL \cite{grill_bootstrap_2020} frameworks. 
We follow the same to check the robustness of All4One compared to NNCLR and Barlow Twins.
As it can be seen in Table \ref{augm}, our approach proves to be more robust than Barlow Twins \cite{zbontar_barlow_2021} and NNCLR \cite{dwibedi_little_2021} to augmentation removal. Intuitively, adding a term that uses multiple neighbours increases, even more, the loss function richness and makes the augmentations less relevant. 

\textbf{Number of Transformer Layers.} 
We increase the  transformer encoder complexity by increasing the number of encoder layers (Table ~\ref{tab:encoder_study}). We infer that more complex encoders decrease the overall performance of the model. 
In SSL paradigms, the main goal is to train an encoder or the backbone. 
We hypothesize that simplifying the transformer encoder encourages the backbone to produce more useful features, rather than letting the transformer encoder create them with their feed-forward layers, thus decreasing performance.

\textbf{Number of Neighbours.} 
We study the effect of the neighbours number on All4One (Table ~\ref{tab:neighbour_study}). Similar to previous works \cite{dwibedi_little_2021, koohpayegani_mean_2021}, All4One is robust regarding the number of neighbours extracted, obtaining the best results when 5 of them are used. As we increase the number of neighbours, the contextual information obtained from the Self-Attention operations performed by the transformer encoder may get more sparse, decreasing slightly the model performance while also increasing the required computation time.

\subsection{Discussions}


\textbf{Dimensionality of the Embeddings.} 
Feature contrast approaches such as Barlow Twins \cite{zbontar_barlow_2021} state high dimensional embeddings as a requirement for their frameworks. However, we prove that combining the basic ideas behind feature contrast with another instance discrimination strategy drastically reduces this dependency. Usually, different SSL approaches tend to be applied individually and obtain improvements. We show that by combining and complementing these ideas could lead to higher improvements. 

\begin{figure}[h!]
\centering
\includegraphics[width=1.0\linewidth,height=5cm]{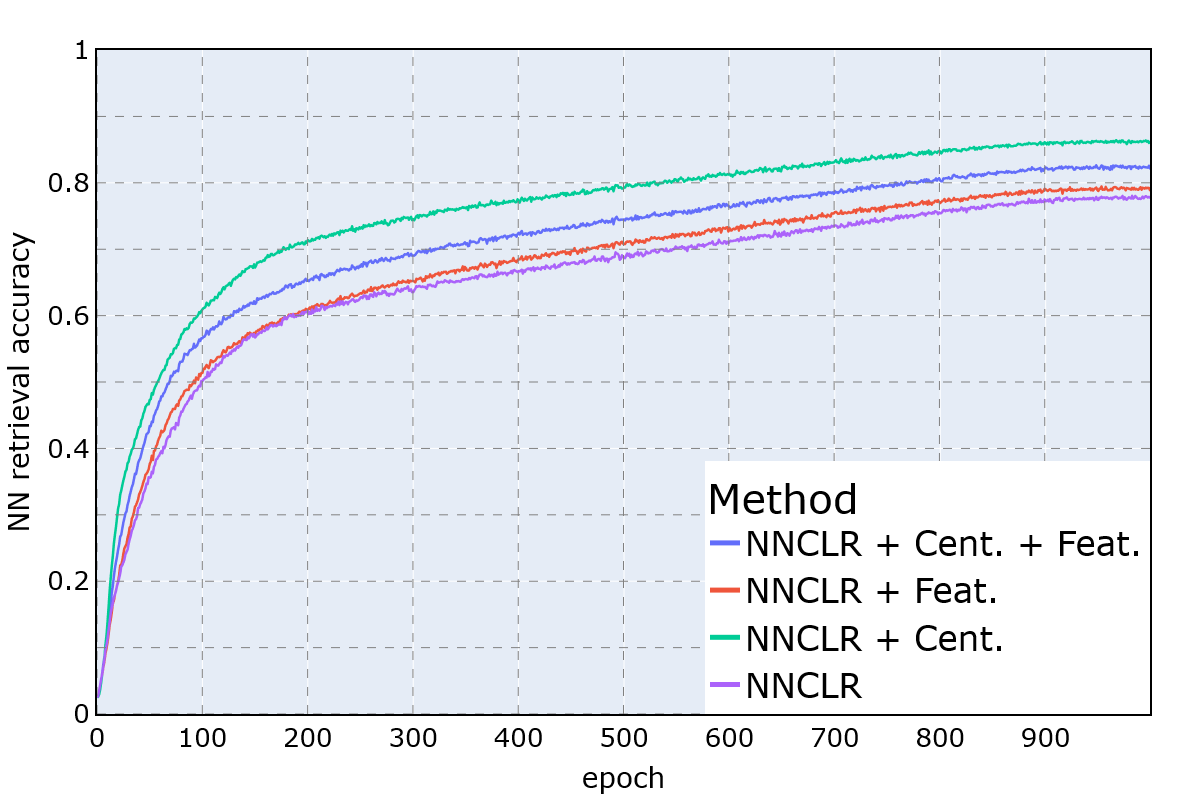}
\caption{\textbf{Top-1 NN retrieval accuracy comparison.}}
\label{NN_acc}
\vspace*{-5mm}
\end{figure}

\begin{figure}[t!]
\centering
\includegraphics[width=.9\linewidth]{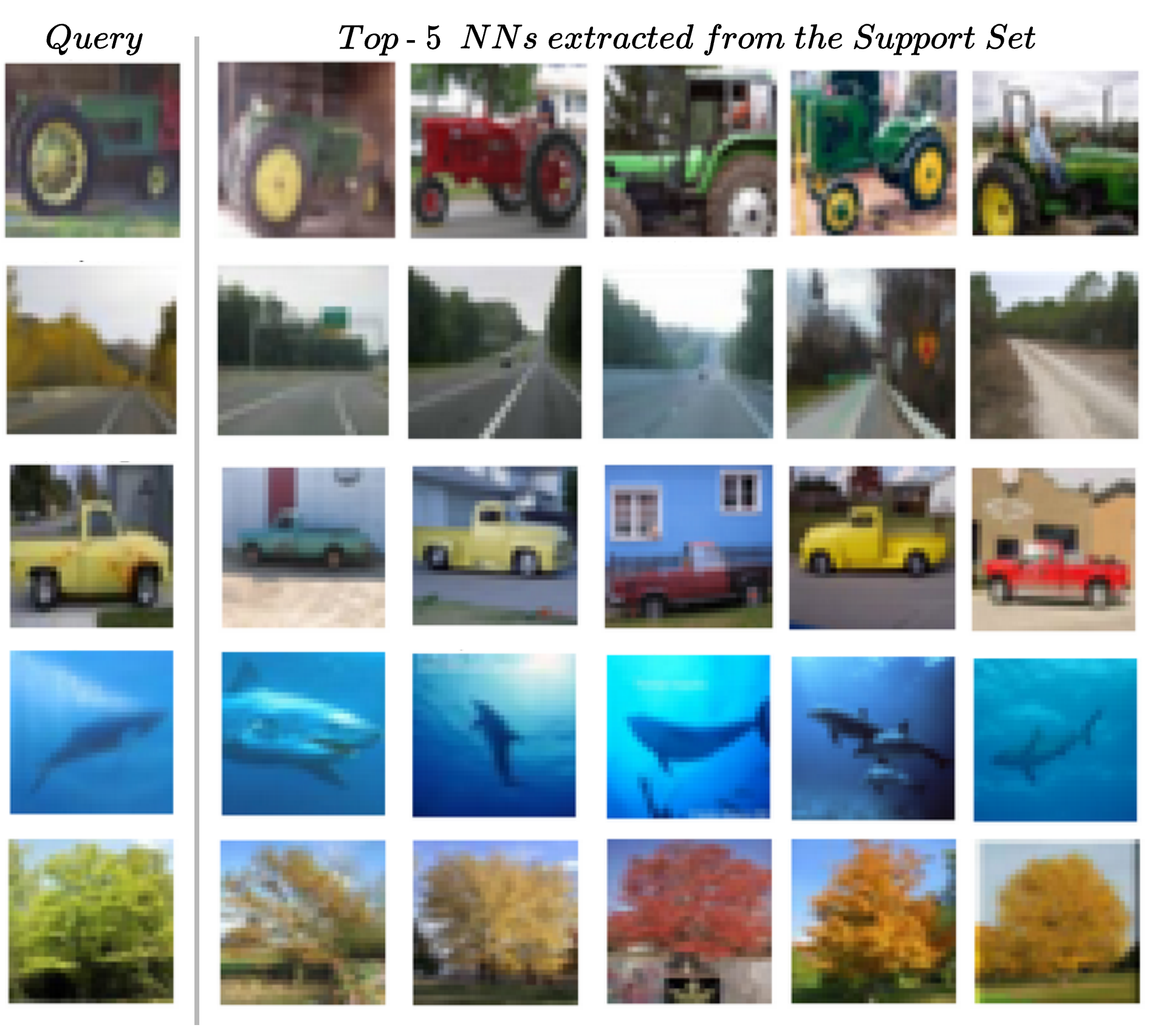}
\caption{\textbf{NN extractions performed by All4One.} More examples in Appendix D}
\label{NN_vis}
\vspace*{-5mm}
\end{figure}

\textbf{NN Retrieval Increase and Comparison.} 
In neighbour contrast approaches, the number of times the KNN operator retrieves a neighbour from the same semantic class as the query (NN retrieval accuracy) has been defined as critical. 
However, we prove that increasing this accuracy does not imperatively lead to a high overall accuracy increase. 
In fact, the best-performing version of All4One is not the one that obtains the highest NN retrieval accuracy. 
Hyperparameters such as the Support Set size or even the pretext task defined affect directly this accuracy. 
In Figure \ref{NN_acc}, we show how adding different pretext tasks affects the original NNCLR \cite{dwibedi_little_2021} method regarding the NN retrieval accuracy. As can be seen, adding the feature contrast strategy to vanilla NNCLR slightly boosts this performance. Also, combining NNCLR with a centroid pretext task provides a NN retrieval accuracy of 86\%. 
However, when the three of them are combined, the NN retrieval accuracy does not surpass the 86\% mark, even if it is the best overall performing version of All4One.
This shows that the NN retrieval accuracy is not as critical as it is stated. 
In fact, as can be seen in Figure \ref{NN_vis}, neighbour contrast frameworks aim to bring together images that are similar so, even if the retrieved neighbour does not belong to the same semantic class, bringing it together with a very similar image on the feature space would not bring down the overall performance of the model.


\subsection{Limitations}
Even though All4One produced promising results, we identify some limitations. 

\textbf{Computation Efficiency.} All4One, due to the introduction of three different objective functions, is more efficient than NN approaches that use multiple neighbours, but less efficient than NNCLR \cite{dwibedi_little_2021}, which only contrasts a single neighbour. This is a limitation on low computation constraints.

\textbf{Increased Number of Hyperparameters.} The final All4One objective function introduces 2 additional hyperparameters to tune. Also, the transformer encoder
uses a different learning rate, which also adds an extra hyperparameter to the overall framework. 

\textbf{Transformer Encoder Parameters.} Several advances are available in terms of training transformers. The stability of the transformer encoder when using different settings compared to that of the other components is not known.

\section{Conclusions}
We propose a symbiotic approach that leverages NN contrastive learning by contrasting contextual information from multiple neighbours in an efficient way via self-attention while also integrating a feature contrast objective function beneficial to the overall framework. 
All4One proves to generalize better and provide richer representations, outperforming previous SoTA contrastive approaches thanks to the integration of its different objectives.
This highlights its exceptional performance in low data regimes, low dimensionality scenarios and weak augmentation settings. 
In future, we plan to extend All4One to more complex backbones and investigate its application in diverse downstream tasks such as Object Detection and Instance Segmentation.

\bibliographystyle{abbrv}
\bibliography{ultra_bibliography}

\section*{Appendix}
\section*{A. All4One Framework}
\textbf{All4One Architecture.} Figure \ref{fig:arch} illustrates a more detailed and comprehensive architecture of the All4One framework, which is designed to accommodate different objective functions for feature, centroid, and neighbour contrast. For the feature contrast objective, image representations generated by the encoder are passed through a projector before the objective function is computed. In the case of the centroid and neighbour objectives, these representations are also passed through a predictor component. Additionally, Centroid path involves extra transformations, owing to the introduction of the transformer encoder and the Shift operation.

\begin{figure*}[t]
\begin{center}
\hspace{1.5cm}\includegraphics[width=0.95\linewidth]{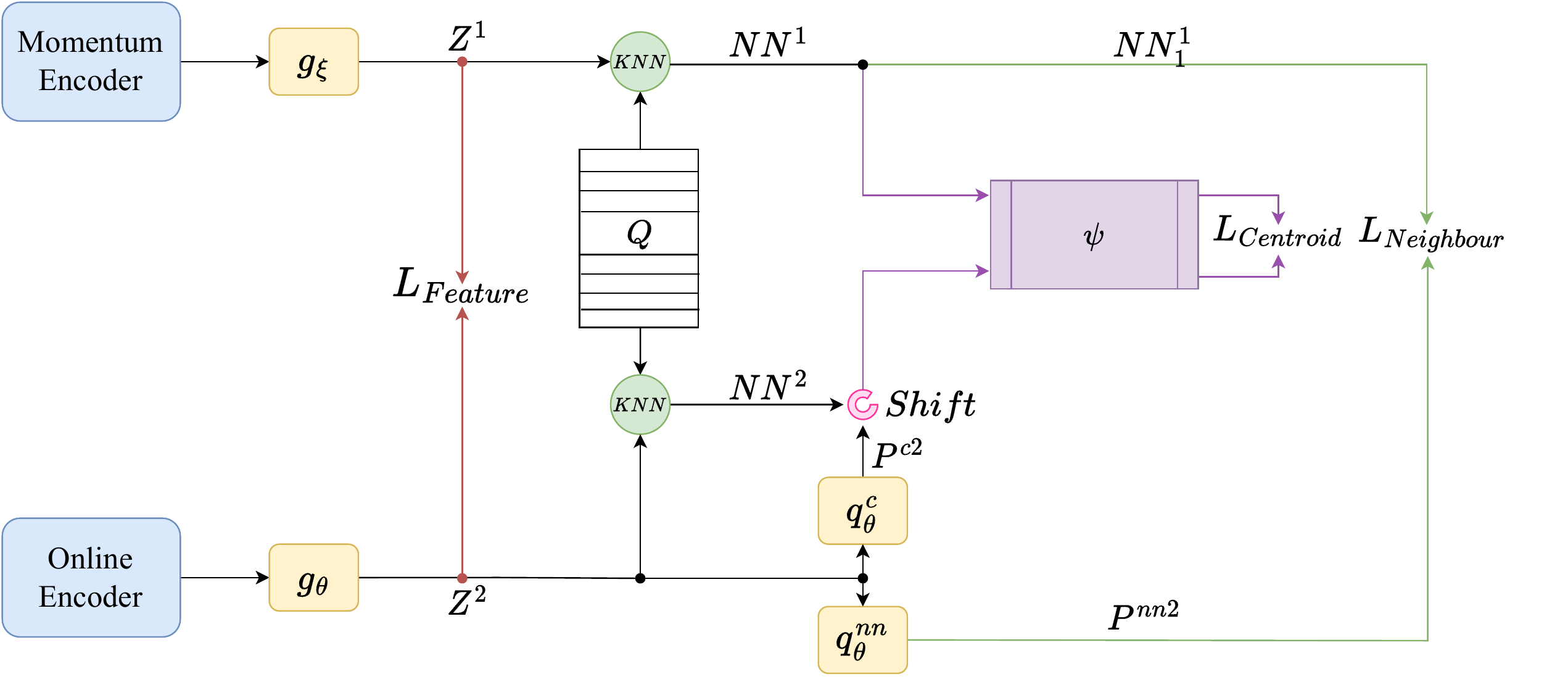}
\end{center}
\caption{\textbf{Complete architecture of All4One framework.} Feature, Centroid and Neighbour contrast objective functions are indicated by red, purple, and green respectively.
}
\label{fig:arch}
\end{figure*}

\begin{figure}[h!]
\begin{center}
\includegraphics[width=0.9\linewidth]{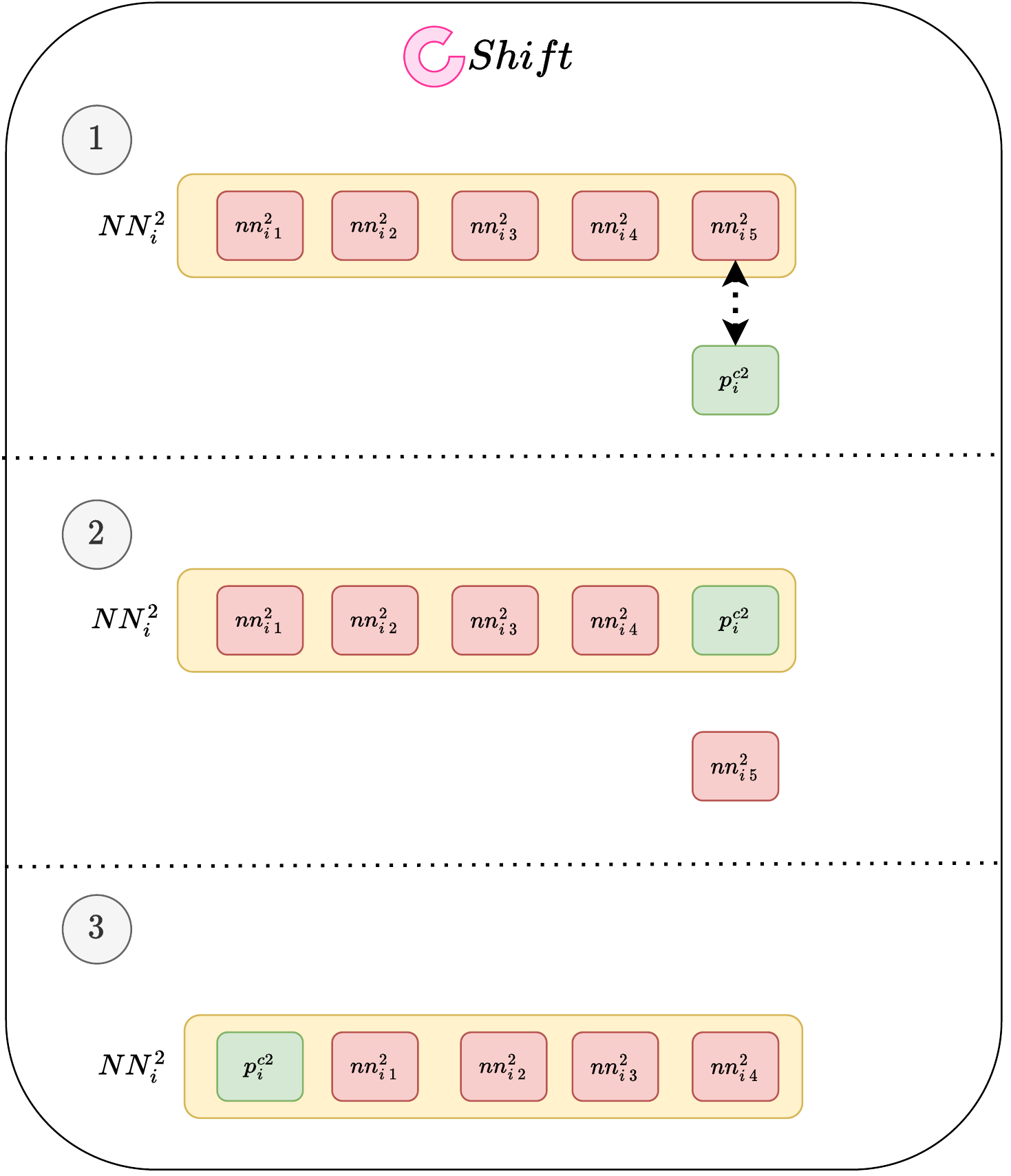}
\end{center}
   \caption{\textbf{Overview of the Shift operation.}}
\label{fig:shift}
\end{figure}

\textbf{Shift operation.} We define the process of replacing an element of a neighbour sequence as Shift operation. This operation can be visualized in Figure \ref{fig:shift}, which illustrates how the final element of a sequence of neighbours is replaced with the output of a predictor, denoted by \(p^{c2}_i\). The resulting sequence is then rearranged so that \(p^{c2}_i\) becomes the first element in the sequence. Note that this operation can be applied to sequences of any length.

\section*{B. UMAP Visualization}
\textbf{Epochwise UMAP for All4One.} In Figure \ref{fig:umap} (left), the progress of five CIFAR-100 classes during training is visually depicted. Initially, the feature representations of these classes lack any structure or pattern, appearing highly mixed. However, as the training process proceeds, the model acquires the ability to generate unique features for each class, causing multiple clusters to emerge within the feature space. These clusters effectively group the feature representations according to their respective classes, even in the absence of annotations, effectively separating one class's representations from another's.

\textbf{Feature representation comparison}. All4One outperforms NNCLR in generating feature representations, enabling more effective cluster visualization by UMAP.  This superior performance is particularly noteworthy in scenarios featuring classes with high similarity, such as those pertaining to certain animal or tree species. As illustrated in Figure \ref{fig:umap} (right), the feature clusters generated from All4One representations for the Maple, Palm, and Willow tree classes demonstrate greater compactness in comparison to those derived from NNCLR representations. Moreover, while the NNCLR approach erroneously groups the Fox and Wolf classes, All4One method distinguishes between these two classes with precision.

\begin{figure}
\centering
\par%
\begin{tabular}{cc}
\multirow{-10}{*}{(a)} & \includegraphics[width=.9\linewidth]{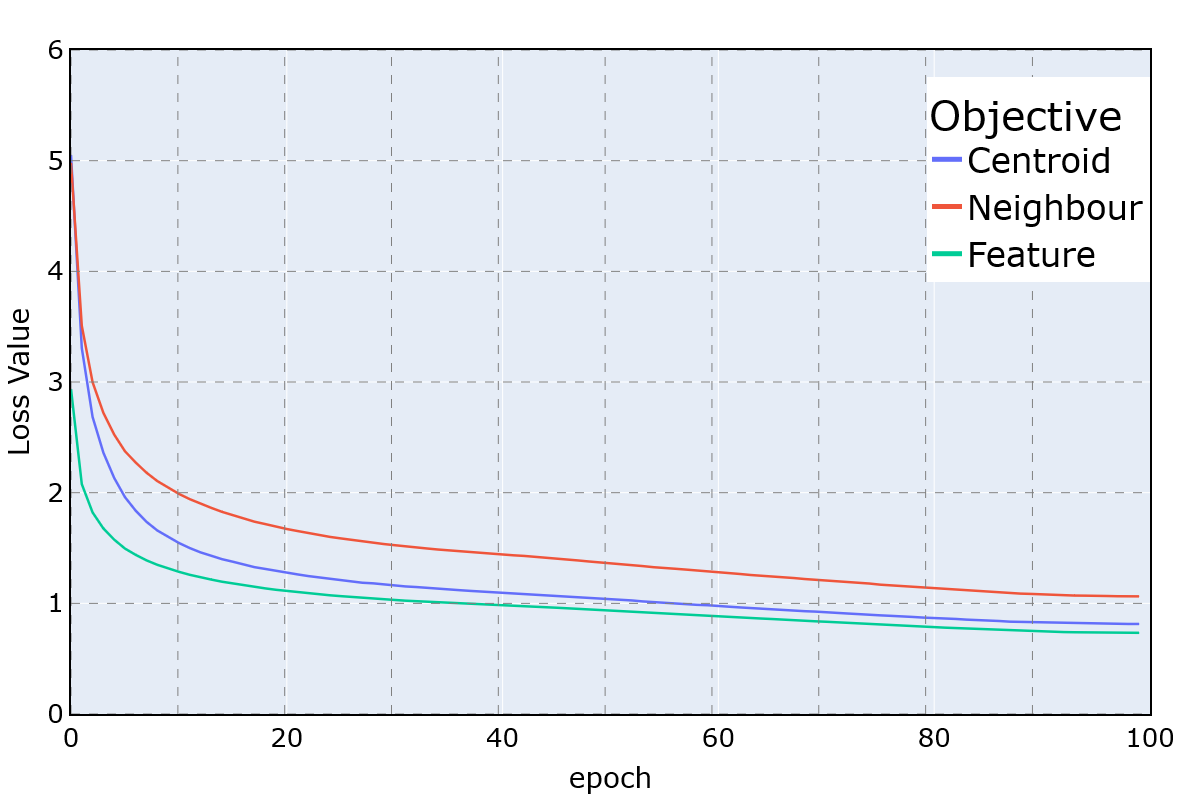}\label{fig:figure_a}\\
\multirow{-10}{*}{(b)} & \includegraphics[width=.9\linewidth]{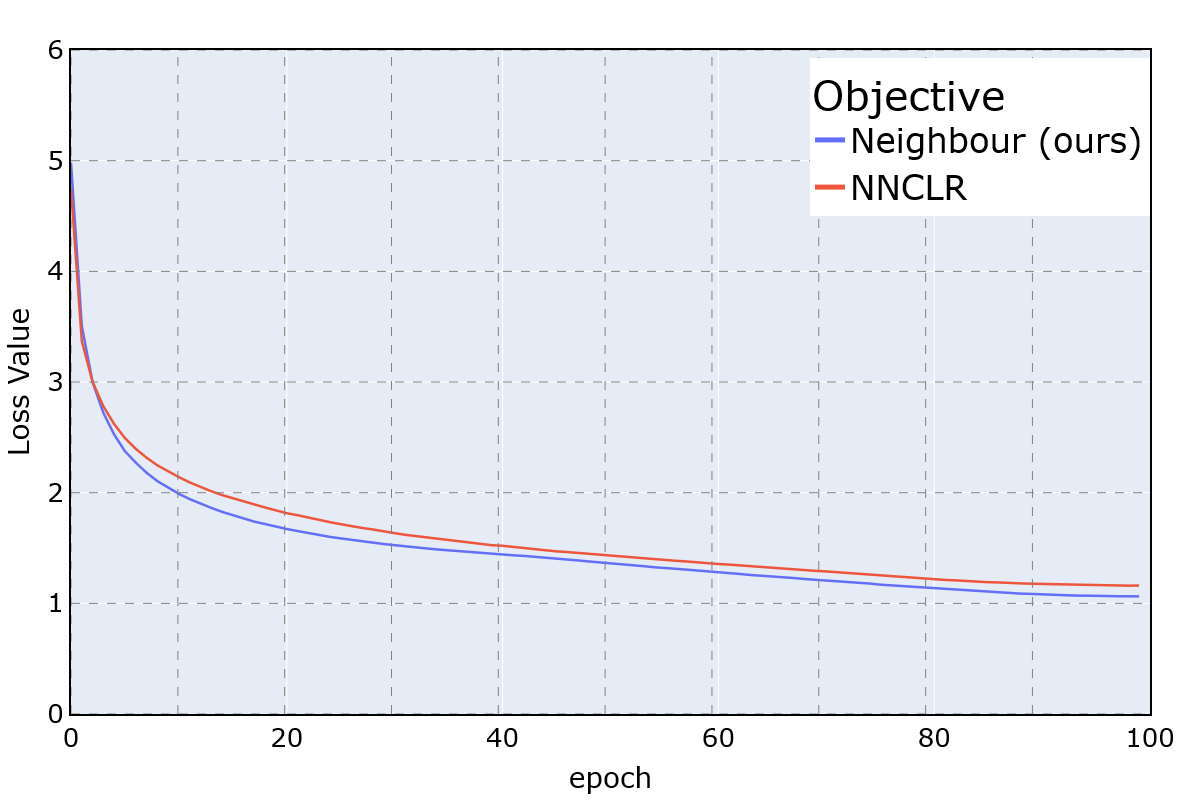}\label{fig:figure_a}\\
\end{tabular}
\caption{\textbf{Training curve comparison (ImageNet). Top:} Centroid vs Neighbour vs Feature. \textbf{Bottom:} NNCLR objective vs All4One Neighbour objective.}.\label{fig:fig1}
\end{figure}

\begin{figure*}
\centering
\includegraphics[width=1.0\linewidth]{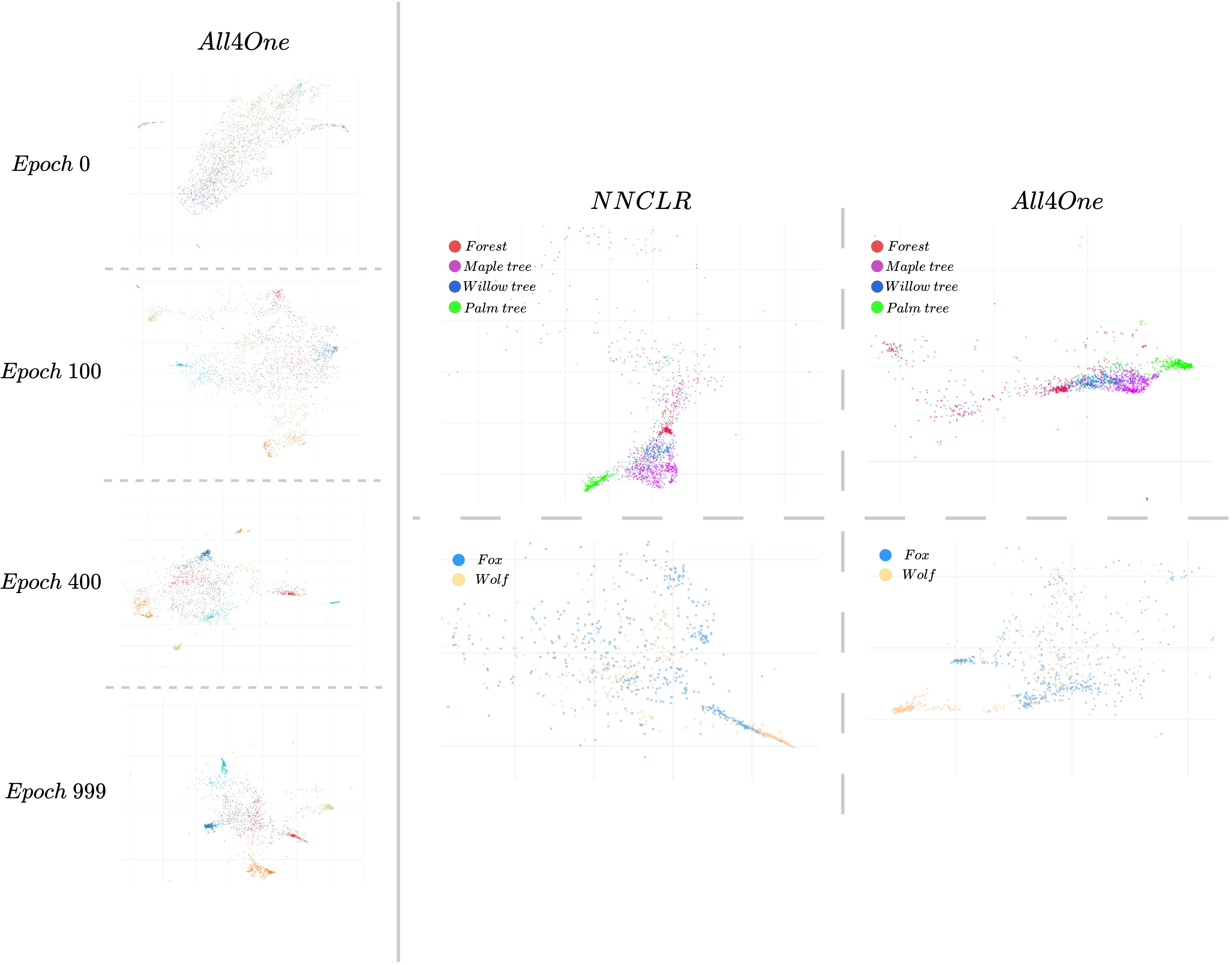}
\caption{\textbf{Left: Epochwise All4One UMAP visualization:} During the training, we select and visualize 5 random classes from CIFAR-100 dataset. \textbf{Right: Per-class UMAP comparison between NNCLR and All4One:} Classes are extracted from CIFAR-100 dataset.
}
\label{fig:umap}
\end{figure*}

\section*{C. Training curves}
\textbf{Feature vs Centroid vs Neighbour.} In Figure \ref{fig:fig1}\textcolor{black}{a}, the training curves for each objective of our model are presented. It is worth noting that the Centroid and Neighbour objectives exhibit very similar starting values, but the Centroid objective is optimized more effectively by our model. We hypothesize that the ease of optimization can be attributed to the contextual information captured by the multiple representations used by the Centroid objective.

\textbf{NNCLR vs Neighbour objective.} Despite the equivalence of the NNCLR objective function and our Neighbour objective function, All4One is able to optimize the latter more effectively, thanks to the symbiotic relationship between the two components, as demonstrated in Figure \ref{fig:fig1}\textcolor{black}{b}, where our Neighbour objective curve is better optimized during training that NNCLR curve.

\section*{D. Neighbour Retrieval}
\begin{figure*}
\centering
\includegraphics[width=0.80\linewidth]{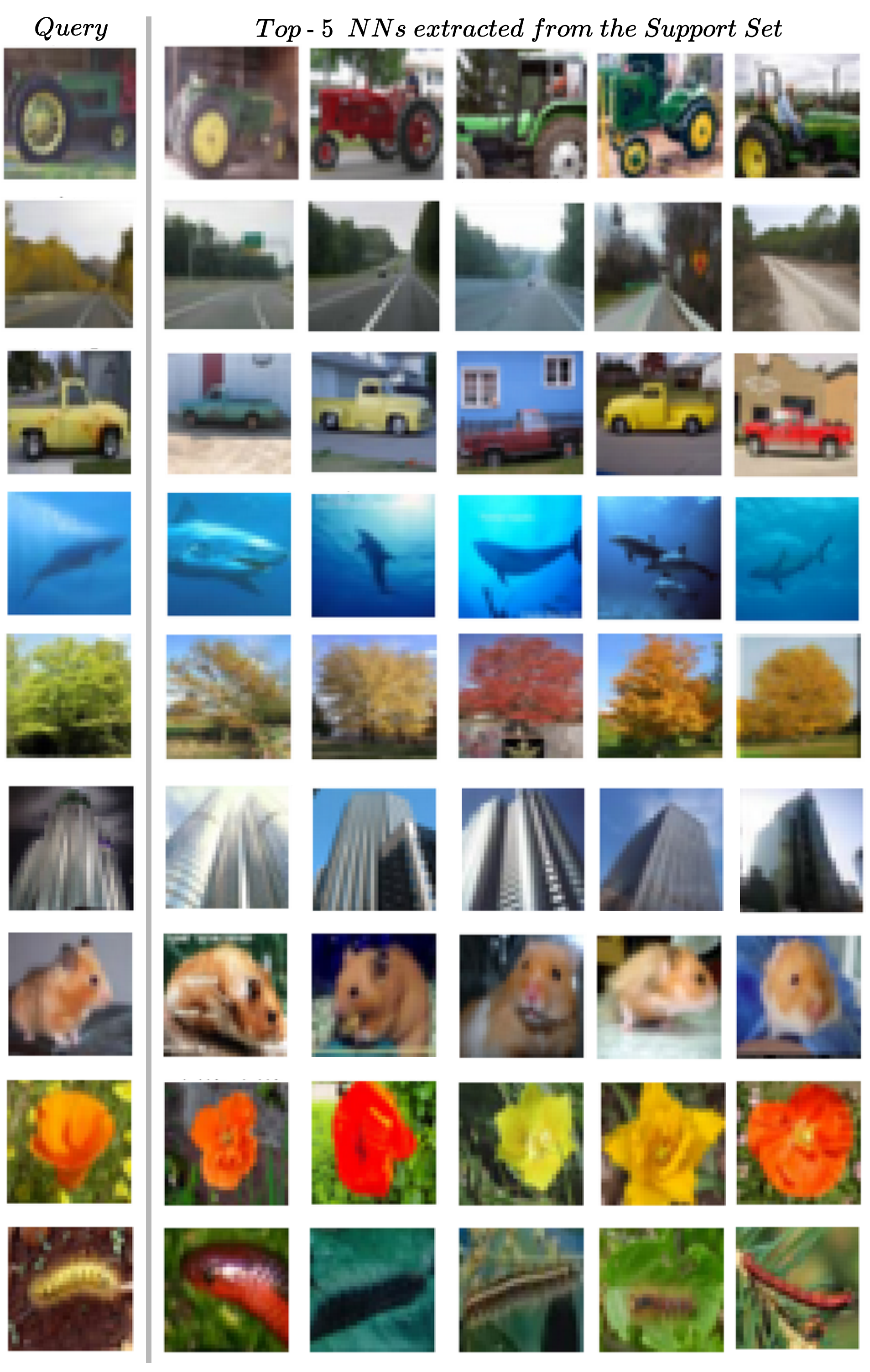}
\caption{\textbf{NN extractions performed by All4One.}}
\label{fig:nn_ret}
\end{figure*}

Figure \ref{fig:nn_ret} showcases some of the image extractions performed by the KNN operator during the training phase. As can be seen, the KNN operator accurately supplies the model with numerous and varied image representations primarily pertaining to the same semantic class.
\end{document}